\documentclass[acmlarge]{acmart-ewaf}

\renewcommand{\baselinestretch}{1.15}

\usepackage{xcolor}
\usepackage{stix} 

\newcommand{\colorsquare}[1]{\textcolor{#1}{$\mdblksquare$}}

\definecolor{c_female}{RGB}{250, 252, 108}
\definecolor{c_male}{RGB}{48, 242, 49}
\definecolor{c3}{RGB}{111, 248, 244}

\AtBeginDocument{%
  }

\setcopyright{cc}
\copyrightyear{2025}
\acmYear{2025}
\acmConference[EWAF'25]{Fourth European Workshop on Algorithmic Fairness}{June 30--July 02, 2025}{Eindhoven, NL}

\begin{document}

\title{When Algorithms Play Favorites: Lookism in the Generation and Perception of Faces}

\author{Miriam Doh}
\affiliation{%
  \institution{ISIA Lab - Université de Mons, IRIDIA Lab - Université Libre de Bruxelles}
  \city{Brussels}
  \country{Belgium}
}
\email{miriam.doh@umons.ac.be}

\author{Aditya Gulati}
\affiliation{%
  \institution{ELLIS Alicante}
  \city{Alicante}
  \country{Spain}}
\email{aditya@ellisalicante.org}
\author{Matei Mancas}
\affiliation{%
  \institution{ISIA Lab - Université de Mons}
  \city{Mons}
  \country{Belgium}}
\email{matei.mancas@umons.ac.be}
\author{Nuria Oliver}
\affiliation{%
  \institution{ELLIS Alicante}
  \city{Alicante}
  \country{Spain}}
\email{nuria@ellisalicante.org}

\renewcommand{\shortauthors}{Doh et al.}

\begin{abstract}
This paper examines how synthetically generated faces and machine learning-based gender classification algorithms are affected by \emph{algorithmic lookism}, the preferential treatment based on appearance. In experiments with 13,200 synthetically generated faces, we find that: (1) text-to-image (T2I) systems tend to associate facial attractiveness to unrelated positive traits like intelligence and trustworthiness; and (2) gender classification models exhibit higher error rates on ``less-attractive'' faces, especially among non-White women. These result raise fairness concerns regarding digital identity systems.
\end{abstract}



\maketitle

\section{Introduction}

Generative Artificial Intelligence (AI) systems are increasingly shaping the content that we consume online \cite{ricker2024ai,Karagianni14112024}. Thus, there is a growing need to detect, quantify and mitigate potential biases that such systems may perpetuate or even amplify \cite{Hall2022}. Although significant work in the literature has focused on gender \cite{Wang2019, Wang2020, Schwemmer2020}, racial \cite{Yucer2020,Howard2024,Khan2021} and age \cite{Karkkainen2021,JacquesJunior2019} biases in computer vision and image generation models, there is growing awareness of the existence of subtler biases \cite{Kumar2024}, such as \emph{lookism}, \emph{i.e.}, the preferential treatment of individuals based on their physical appearance. Rooted in beauty standards and cognitive biases \cite{Dion1972, Talamas2016, Eagly1991, Tversky1974, gulati2024beautifulgoodattractivenesshalo}, lookism can lead to systemic disadvantages and discrimination for individuals who do not conform to prevailing aesthetic norms, affecting their opportunities and how they are perceived and judged by automated AI systems. 

Existing evaluations of text-to-image (T2I) models have revealed demographic biases, with whiteness and masculinity overrepresented in the generated images \cite{25_luccioni2023stable,26_naik2023social}. These biases extend beyond demographic traits, affecting object selection, clothing, and even spatial representations \cite{wu2023stable}.

In this preliminary study, we extend the evaluation of text-to-image (T2I) models to the domain of algorithmic lookism \cite{gulati2024lookism}, examining the relationship between facial attractiveness and other behavioral traits in images of faces generated using Stable Diffusion 2.1 \cite{rombach2022highresolutionimagesynthesislatent}. Specifically, we assess four traits—happiness, sociability, trustworthiness, and intelligence—originally operationalized by Gulati et al. \cite{gulati2024beautifulgoodattractivenesshalo} in their large-scale study of the attractiveness halo effect through the application of beauty filters to pictures of human participants.

Our selection of these traits is grounded in a longstanding body of social-psychological literature (\cite{Dion1972,Kanazawa2004,Talamas2016,Mathes1975,Golle2013,Todorov2008,Miller1970}), which has consistently demonstrated that physically attractive individuals are \textit{perceived} to be more 
happy, sociable, trustworthy, and intelligent. By adopting the same trait definitions utilized by Gulati et al., our study facilitates direct comparability between results obtained from synthetic facial images and the large-scale, high-quality human data reported in their work, enhancing potential for future research.

Furthermore, we evaluate the impact of algorithmic lookism on three gender classification models applied to synthetically generated face images with different attributes. This analysis builds on work by Doh et al. \cite{doh2024my} which showed that gender classification models perform better on \emph{attractive} AI-generated female faces.

As a result of our study, we obtain three key findings:
(1) T2I generative models show strong signs of algorithmic lookism, associating attractiveness with positive behavioral traits, even though attractiveness is not necessarily a good predictor of such traits;
(2) Gender classification models are also impacted by algorithmic lookism, with faces generated using negative trait terms being misclassified more often than their positive counterparts; and
(3)	The impact of algorithmic lookism is non-uniform across gender and racial groups with Asian and Black women being disproportionately impacted. 

 \section{Dataset Creation and Methodology}
 \label{sec:dataset}
A total of 13,200 face images were generated using Stable Diffusion 2.1 \cite{rombach2022highresolutionimagesynthesislatent}, varying gender (woman, man), race\footnote{The term \textit{race} is used as it appears in standard ML datasets, acknowledging it as social construct distinct from \textit{ethnicity} \cite{APA_Race,APA_Ethnicity}. This study does not seek to promote the use of racial categories in AI, nor do we aim to reify race through generative AI. Instead, our objective is to critically examine how AI systems encode and propagate biases, including those linked to socially constructed categories like race} (Asian, Black, White), and five attribute pairs associated with the attractiveness halo effect \cite{gulati2024beautifulgoodattractivenesshalo}: attractive vs. unattractive, intelligent vs. unintelligent, trustworthy vs. untrustworthy, sociable vs. unsociable, and happy vs. unhappy. 

\begin{figure}[ht] \centering \includegraphics[width=\linewidth]{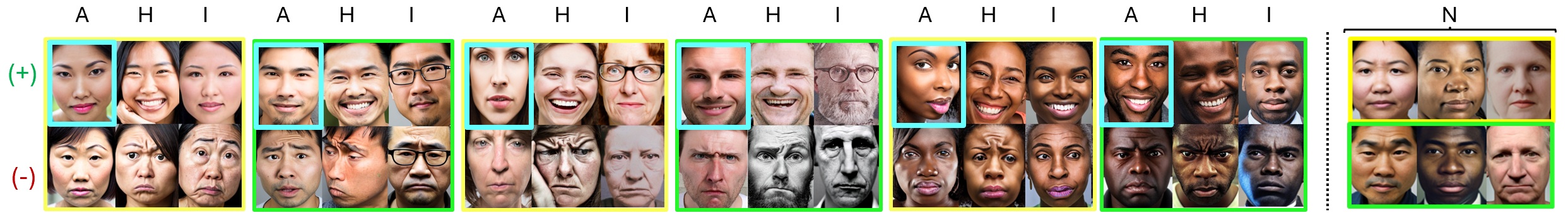} \caption{Examples of the generated faces with Stable Diffusion 2.1 with positive (+) and negative (-) variations for three traits (A = Attractiveness, H = Happiness, and I = Intelligence) together with the neutral faces (N = Neutral). Yellow (\colorsquare{c_female}) and green (\colorsquare{c_male}) correspond to images of females and males, respectively. Light Blue (\colorsquare{c3}) borders highlight the faces corresponding to the positive Attractiveness trait.} \label{fig:generared_faces} \end{figure}

Prompts followed the format ``\texttt{Front photo of a [attribute] [race] [gender]}'', with 200 images per triplet. Additionally, we generated a ``Neutral'' set (200 images per combination) without attribute descriptors, serving as a baseline to assess the model’s default tendencies.
Figure \ref{fig:generared_faces} shows samples of the images for positive and negative trait terms across all six gender and race categories.

We extracted the CLIP embeddings \cite{radford2021learningtransferablevisualmodels} of the generated images to quantify the relationship between perceived attractiveness and other facial traits. Specifically, for each triplet \([\mathrm{attribute},\mathrm{race},\mathrm{gender}]\) we collect \(N\) 512-dimensional vectors \(\{\mathbf{e}_i\}_{i=1}^N\subset\mathbb{R}^{512}\) and compute their centroid 
\(\mathbf{c} = \frac{1}{N}\sum_{i=1}^N \mathbf{e}_i\).  
We then measure the Euclidean distance \(d(\mathbf{c}_a,\mathbf{c}_b)=\|\mathbf{c}_a-\mathbf{c}_b\|_2\) between any two centroids \(\mathbf{c}_a,\mathbf{c}_b\) and define the similarity score as \(\mathit{sim}=1/d(\mathbf{c}_a,\mathbf{c}_b)\), since smaller distances correspond to a higher similarity.

Similarity scores were computed between images generated using positive trait descriptors and the `attractive' faces as well as between negative trait descriptors and the `unattractive' faces. A two-sided t-test was conducted to assess the statistical significance of the centroid distance computed.

Gender classification was carried out using three popular models: InsightFace \cite{ren2023pbidr}, DeepFace \cite{serengil2021lightface}, and FairFace \cite{karkkainenfairface}. Accuracy was tested across all positive and negative traits, gender, and race, and misclassification rates were analyzed to examine potential differences in classification errors across demographic and attractiveness categories.

Note that in this research we do not define or measure attractiveness, but focus on analyzing how T2I models associate attractiveness, or it's lack thereof, with other positive and negative attributes.

\section{Results}
\label{sec:results}
\begin{figure}[h]
    \centering
    \includegraphics[width=\linewidth]{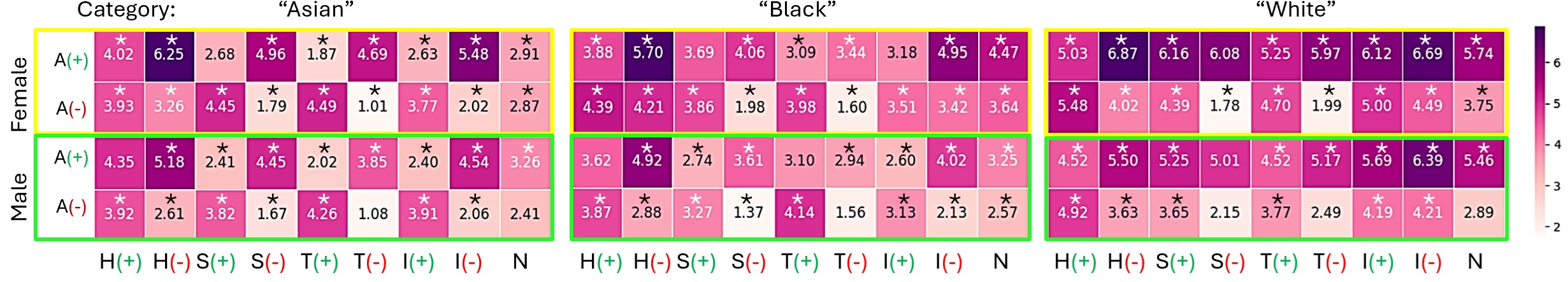}
    \caption{Heatmaps of centroid distances between attractiveness (A) and other social traits (happiness (H), sociability (S), trustworthiness (T), and intelligence (I)), across gender and racial groups. Lower centroid distances values indicate stronger associations. Positive (+) and negative (-) variations represent trait polarities. Significant results ($p < 0.05$) are marked with *}
    \label{fig:centroid_heatmap}
\end{figure}

\textbf{Lookism in T2I Models:} Each cell in Figure \ref{fig:centroid_heatmap} represents the computed distance between two image groups, as specified by the corresponding row and column. T-tests were carried out to confirm that the distributions are statistically different ($p < 0.05$). As seen in the figure, the centroids of the images generated with positive trait terms tend to be closer to images of attractive individuals while images generated with negative trait terms tend to be closer to images of unattractive individuals. This pattern, however, varies across gender and race.

The faces corresponding to Asian and Black women exhibit a strong alignment between attractiveness with positive traits, and unattractiveness with negative traits. Hence, we observe the existence of \emph{lookism} in these cases. 

Interestingly, the faces of White women show weaker or inconsistent associations, such that faces created with positive traits are closer to unattractive rather than attractive faces. Among men, the trend is similar but less pronounced than for women, with faces of Asian men showing the strongest lookism. Similar to what we observe with faces of White women, the faces of White men also exhibit inconsistent associations across some attributes. Neutral faces are generally closer to unattractive rather than attractive faces across all demographics, with the effect most pronounced for White individuals.

\begin{figure}
    \centering
    \includegraphics[width=\linewidth]{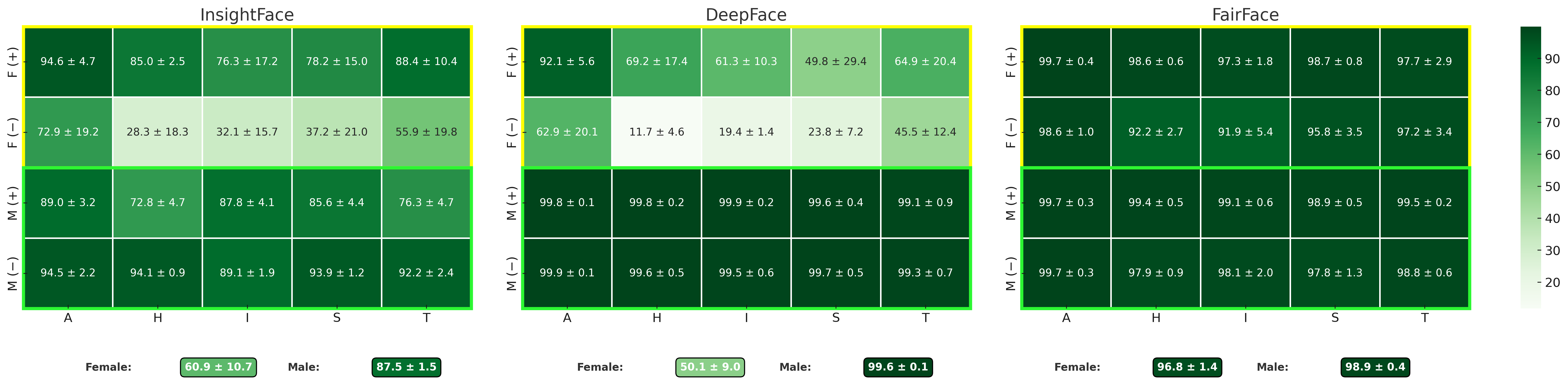}
    \caption{Heatmaps of gender classification accuracy (Mean ± Std) for InsightFace, DeepFace, and FairFace. A = Attractiveness, H = Happiness, I = Intelligence, S = Sociability, T = Trustworthiness. Female = Yellow \colorsquare{c_female}, Male = Green \colorsquare{c_male}. The two values below each heatmap represent the classification accuracy for neutral female and male faces.}
    \label{fig:classification}
\end{figure}

\textbf{Gender Classification:} Figure \ref{fig:classification} shows the varying performance of the three different gender classification models across gender and attribute groups. 

The classification accuracy of InsightFace is high for attractive female faces (\textbf{94.6\%} ± 4.7), with all positively associated traits remaining above 76\% with this model. However, the same model exhibits much lower accuracy on unattractive female faces (73.0\% ± 19.1), with significant accuracy drops for certain negatively associated traits.

Interestingly, classification accuracy remains high for male faces across all traits ($\approx\mathbf{85\%}$), and faces generated with negative attributes are classified with greater accuracy contrary to the observed trend for female faces.

DeepFace achieves near-perfect accuracy on male faces across both positive and negative attractiveness values and traits, while the same model yields much worse accuracies on female faces, especially those created with negative traits. Particularly poor is the performance on the faces of generated with the negative attributes unhappy (11.7\% ± 4.6) and unsociable (19.5\% ± 7.1) for women. Notably, the only trait for which the model achieves above 90\% accuracy on female faces is Attractive (+), whereas the performance on faces created with all other attributes—both positive and negative—is significantly worse compared to the accuracy obtained on images of males.

FairFace's performance is the most balanced and competitive across genders with accuracies above 90\% in all cases. While the gap in performance between the faces created with positive and negative traits is smaller than for InsightFace and DeepFace, images of women still experience worse classification results than images of men, especially when created with negative attributes—\emph{e.g}., Happy (-) (92.2\% ± 2.7) vs Happy (+) (98.6\% ± 0.6) and Intelligent (-) (91.9\% ± 5.4) vs Intelligent (+) (97.3\% ± 1.8), though less extreme than in other models.

These preliminary findings suggest that algorithmic lookism in T2I models affects the performance of gender classification systems, with images of females being more notably impacted than images of males. However, further investigation is necessary to explore the full extent of this effect and to identify the underlying causes for the differential impact observed across the three models examined in this study.

\section{Discussion and Conclusion}
\label{sec:discussion}
This study provides evidence of algorithmic lookism in T2I models, where attractiveness is linked to positive traits, particularly for Asian and Black women. White faces, by contrast, exhibit greater visual diversity, suggesting possible dataset limitations for the other race categories that warrant further investigation. Regarding gender classification, our finding aligns with prior research \cite{buolamwini2018gender,doh2024my}, showing higher misclassification rates for women. One plausible interpretation of the markedly worse performance on “negative” female faces is that the generative model yields images with noticeably different visual cues in those scenarios. 

Doh et al. \cite{doh2024my} demonstrated that ``unattractive'' synthetic female faces tend to appear older, display neutral or downward‐turned expressions,  and—particularly for Asian faces—lack any makeup; notably, we observe these very same characteristics consistently across most of our negative‐trait categories (Figure \ref{fig:generared_faces}).
In particular, the possible impact of the absence of makeup aligns with Muthukumar et al.\cite{muthukumar2018understandingunequalgenderclassification}'s finding that state-of-the-art gender classifiers rely heavily on cosmetic features around the eyes and lips when identifying ``female'', whereas male classification uses different cues. If the GenAI pipeline omits or diminishes these makeup regions, the classifier loses a key signal and errs more frequently.

This highlights a broader issue: algorithmic systems do not merely misclassify; they shape visibility itself. As Butler's concept of gender intelligibility \cite{butler2011bodies} and De Lauretis’ \textit{technologies of gender} \cite{deLauretis_1987} suggest, AI-generated identities reinforce dominant sociotechnical frameworks. As T2I models integrate into social media \cite{bruns2024you}, advertising, and downstream applications like data augmentation \cite{benkedadra2024cia,chen2024would}, they risk entrenching biases into algorithmic infrastructures.
The present study, while preliminary, yields valuable insights into the influence of lookism on text-to-image (T2I) models. To further this research, we identify four key areas: (1) Disentangling the findings from biases that could potentially originate from the CLIP embeddings (2) assessing the impact of algorithmic lookism on downstream AI applications, particularly classification models trained on synthetic data; (3) investigating whether T2I models encode a standardized concept of attractiveness, shaping their outputs accordingly; and (4) conducting a targeted XAI study to confirm the relative contributions of apparent age, facial expression, and makeup absence to the observed gender‐classification bias.

\begin{acks}
M.D. acknowledges support from the ARIAC project (No. 2010235), funded by the Service Public de Wallonie (SPW Recherche), and funding from the FNRS (National Fund for Scientific Research) for her visiting research at the ELLIS Alicante Foundation. A.G. and N.O. are partially supported by a nominal grant received at the ELLIS Unit Alicante Foundation from the Regional Government of Valencia in Spain (Convenio Singular signed with Generalitat Valenciana, Conselleria de Innovaci\'on, Industria, Comercio y Turismo, Direcci\'on General de Innovaci\'on), along with grants from the European Union’s Horizon 2020 research and innovation programme (ELIAS; grant agreement 101120237) and Intel. A.G. is additionally partially supported by a grant from the Banc Sabadell Foundation. Views and opinions expressed are those of the author(s) only and do not necessarily reflect those of the European Union or the European Health and Digital Executive Agency (HaDEA).  
\end{acks}

\bibliographystyle{ACM-Reference-Format}
\bibliography{sample-base}

\appendix









\end{document}